\documentclass[11pt]{article}
\usepackage{acl}
\usepackage{times}
\usepackage{latexsym}
\usepackage[T1]{fontenc}
\usepackage[utf8]{inputenc}
\usepackage{microtype}
\usepackage{inconsolata}
\usepackage{graphicx}
\usepackage{amsmath}
\usepackage{amssymb}
\usepackage{booktabs}
\usepackage{multirow}
\usepackage{xcolor}
\usepackage{subcaption}
\usepackage{float}
\usepackage{needspace}
\usepackage{makecell}

\newcommand{\roi}[1]{\texttt{<roi\textsubscript{#1}>}}
\newcommand{\obj}[1]{\texttt{<obj\textsubscript{#1}>}}

\title{From Coordinates to Candidate Regions:\\Temporal Change Localization via Region Selection\\in Remote Sensing Multimodal LLMs}

\author{Juwan Chung \quad Sungjune Park \quad Yeongyun Kim \quad Yong Man Ro\thanks{\hspace{1mm}Corresponding author.} \\
  Integrated Vision Language Lab, KAIST, South Korea \\
  \texttt{\{juwan99, sungjune-p, yeongyun.kim, ymro\}@kaist.ac.kr}}

\begin{document}
\maketitle

\begin{abstract}
Remote sensing multimodal large language models (RS-MLLMs) have advanced scene understanding and visual question answering over satellite imagery, yet localizing specific objects or changed regions remains challenging.
Existing approaches rely on generating bounding box coordinates as token sequences, which is fragile for the small, densely packed objects common in remote sensing and increasingly error-prone when multiple targets must be localized simultaneously.
In this work, we present an RS-specific formulation of the region selection paradigm, previously explored in natural-image MLLMs, and extend it to temporal change localization over multi-image sequences.
Our framework employs a text-conditioned region proposal module, encodes each candidate as special tokens carrying per-frame visual features enriched with spatial and temporal cues, and lets the LLM localize targets by selecting region tokens in its response.
We construct a multi-task training and evaluation suite spanning localization, referring expression, visual grounding, and understanding tasks across single-image and multi-temporal settings.
Experiments show that our approach substantially outperforms coordinate-generation baselines on temporal change localization, while improving single-image visual grounding and maintaining competitive understanding performance.
Oracle analysis decomposes the contributions of the region proposer and the LLM selector, providing diagnostic insight unique to this framework.
Our code will be available at \url{https://github.com/juwan-kr/RS-RegionSelect}.
\end{abstract}

\section{Introduction}
\label{sec:intro}

Remote sensing (RS) image analysis, including object detection, change detection, and scene classification, plays a critical role in disaster response, urban planning, and environmental monitoring.
The recent introduction of multimodal large language models (MLLMs) to this domain has brought significant progress, enabling conversational interaction with RS imagery \cite{geochat,lhrsbot,earthgpt,skyeyegpt,earthdial,park2025remote}, temporal multi-image reasoning \cite{teochat}, and fine-grained spatial grounding at both box and pixel levels \cite{geoground,geopixel,terrascope,shu2025earthmind}.

Despite these advances, \emph{localization} of specific objects or changed regions remains a key challenge.
Most existing RS-MLLMs perform localization by generating bounding box coordinates as discrete token sequences.
This coordinate-generation approach can be fragile in the RS setting, where objects are typically small and densely packed, occupying only a handful of image patches; small coordinate errors that would be negligible for large natural-image objects can significantly degrade localization quality.
The problem is compounded when multiple regions must be localized at once, as in change detection, because the model must produce a long sequence of coordinate tokens with growing risk of omission, duplication, or degenerate repetitive patterns (Figure~\ref{fig:qualitative}).

\begin{figure*}[t]
\centering
\includegraphics[width=\textwidth]{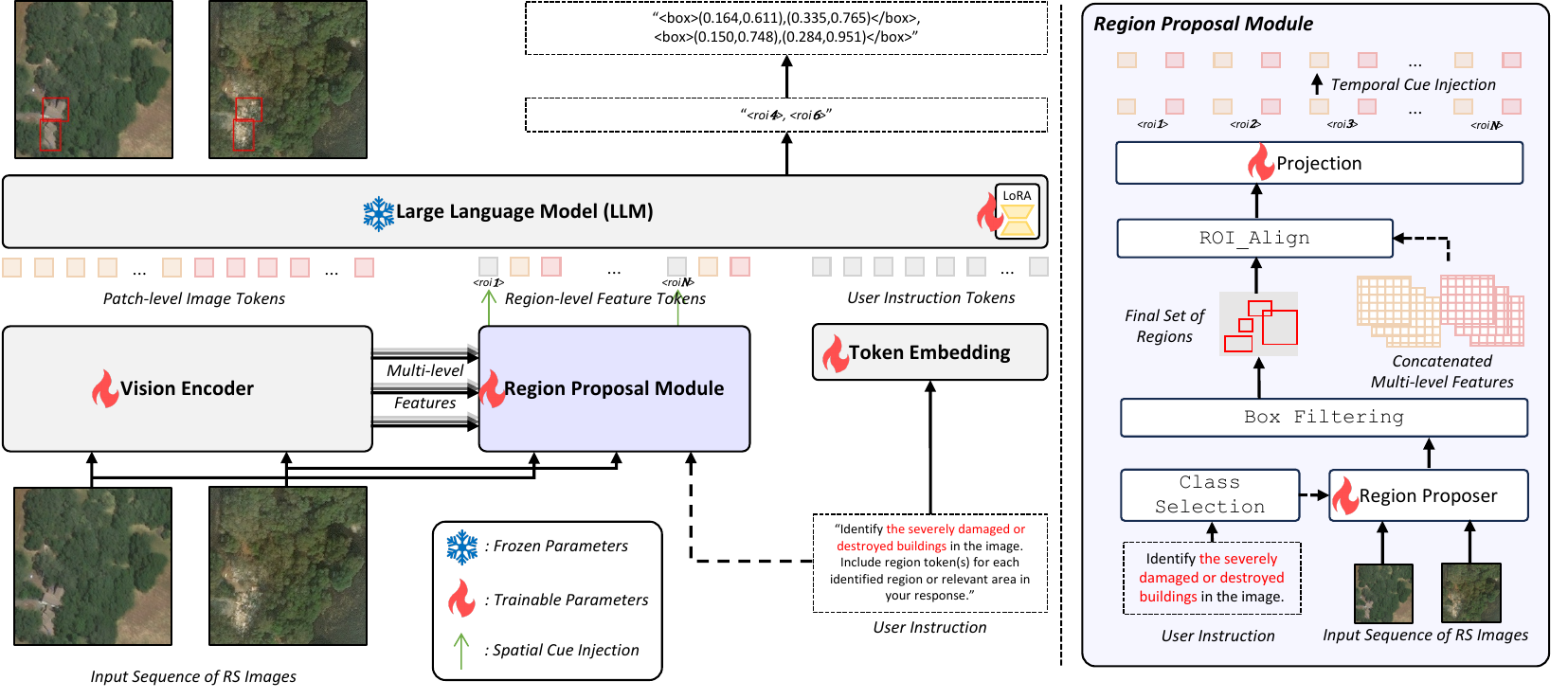}
\caption{Overview of our framework. Given an input sequence of RS images and a user instruction, the vision encoder extracts multi-level features. The region proposal module (right) performs text-conditioned class selection, generates region proposals, applies box filtering and ROIAlign \cite{roialign}, and projects features into region tokens with spatial and temporal cue injection. The LLM receives patch-level image tokens, region-level feature tokens, and user instruction tokens, and generates a response containing selected region tokens (e.g., \roi{4}, \roi{6}) for localization.}
\label{fig:overview}
\end{figure*}

To address this, we draw on recent work in natural-image MLLMs that frames localization as \emph{region selection} \cite{groma,chatrex,gpt4roi,osprey}: given a set of candidate regions, the model selects the relevant ones rather than generating coordinates from scratch.
We adopt this paradigm and extend it to the RS domain, where temporal image sequences and heterogeneous object vocabularies present additional challenges.
While open-vocabulary detection in RS has been explored by standalone detectors \cite{openrsd}, integrating such detection as a component within an MLLM for region-level reasoning over temporal sequences has not been addressed.
Our framework represents each candidate region as a group of special tokens carrying per-frame visual features along with spatial and temporal cues (Figure~\ref{fig:overview}).
The LLM performs localization by generating region tokens within its natural language response, effectively reducing the localization problem to discrete token selection.

To support training and evaluation, we assemble a multi-task dataset from both multi-temporal sources (TEOChatlas \cite{teochat} for change localization and temporal QA; additional change detection datasets \cite{levirmci,tuecd,s2looking,egybcd}) and single-image sources (GeoChat Instruct \cite{geochat}, FIT-RS \cite{fitrs}, DIOR-RSVG \cite{diorrsvg}).
Experiments show that our model substantially outperforms both existing RS-MLLMs and a same-data coordinate-generation baseline on temporal change localization and visual grounding tasks, while maintaining competitive performance on scene classification, QA, and other understanding tasks.

Our contributions are as follows:
\begin{itemize}
\item We present an RS-specific formulation of region selection for dense localization and extend it to temporal change detection over multi-image sequences. Each candidate carries aligned per-frame visual features, enabling region-level cross-frame comparison within the LLM.

\item We design a region proposal module that integrates heterogeneous RS detection datasets into a unified class vocabulary and employs prompt-aware class filtering to bridge the LLM's natural language query to the detector's class space.

\item We construct a multi-task training and evaluation suite and provide systematic evaluation with controlled ablations and oracle analysis that separately quantifies the proposer and selector contributions.
\end{itemize}

\section{Related Work}
\label{sec:related}

\paragraph{Remote Sensing MLLMs and Temporal Reasoning.}
A growing body of work has adapted MLLMs to remote sensing.
GeoChat \cite{geochat} introduced conversational capabilities with region-level referring and grounding.
TEOChat \cite{teochat} extended MLLM reasoning to temporal earth observation, supporting diverse spatio-temporal tasks including change localization, damage assessment, and temporal QA over multi-image sequences.
EarthDial \cite{earthdial} supports both multi-sensor and multi-temporal inputs.
Other systems target pixel-level grounding \cite{geopixel,terrascope}, cross-sensor fusion \cite{shu2025earthmind}, and large-scale scene understanding \cite{lhrsbot,earthgpt,skyeyegpt}.
For language-conditioned change reasoning, ChangeChat \cite{changechat} and CDChat \cite{cdchat} address bi-temporal conversational understanding, while TerraScope \cite{terrascope} integrates pixel-level masks with multi-temporal chain-of-thought reasoning.
Pixel-level approaches offer fine spatial resolution but can struggle to delineate individual objects in densely packed scenes, where instance-level distinctions matter.
Among box-level approaches, localization is universally performed through coordinate generation.
Our work explores region selection as a complementary alternative that aligns with the LLM's discrete token generation mechanism and provides explicit per-region visual features.

\paragraph{Region-Level Visual Grounding in MLLMs.}
Visual grounding in MLLMs is typically performed either by generating bounding box coordinates \cite{shikra,kosmos2,qwen2} or by producing segmentation masks \cite{lisa}.
A complementary line of work injects region-level features directly into the LLM input.
GPT4RoI \cite{gpt4roi} uses ROIAlign features as spatial tokens; Groma \cite{groma} introduces a locate-then-understand pipeline with a dedicated region proposer; ChatRex \cite{chatrex} adds retrieval-based region perception; and Osprey \cite{osprey} supports pixel-level region understanding.
These approaches have proven effective for natural images but are limited to single-image settings.
On the detection side, open-vocabulary object detection has been explored in RS \cite{openrsd}, demonstrating that text-conditioned detectors can generalize across heterogeneous RS class vocabularies.
We build on these foundations and extend the region-centric paradigm to RS with two key adaptations: each candidate carries per-frame visual features from a temporal image sequence for cross-time comparison, and the region proposal module employs prompt-aware class filtering to bridge the LLM's natural language query to the detector's class vocabulary.

\section{Method}
\label{sec:method}

\subsection{Task Formulation}
\label{sec:task}

Region-level candidate selection has been explored in natural-image MLLMs \cite{groma,chatrex,gpt4roi}.
We adopt this paradigm and extend it to RS, where temporal image sequences and dense small objects present distinct challenges.

Given RS images $\mathcal{I} = \{I_1, \ldots, I_M\}$ capturing the same location at different times and a text query $q$, the goal is to identify the spatial regions relevant to $q$.
In the conventional approach, the model generates bounding box coordinates as a token sequence.
In our formulation, a region proposal module first produces a set of candidates $\mathcal{R} = \{r_1, \ldots, r_N\}$, and the model selects relevant ones by generating their corresponding tokens (e.g., \roi{4}) within a natural language response.
Each selection is a single token rather than a multi-token coordinate tuple, which not only simplifies the output space but also reduces the number of tokens the LLM must generate for multi-target localization, improving inference efficiency (Appendix~\ref{app:efficiency}).

Beyond simplifying the output, this formulation also enriches the input: the visual features of each candidate are injected into the LLM's context, giving the model direct access to region-level visual information that is absent in coordinate-generation approaches.

\subsection{Region Proposal Module}
\label{sec:rpm}

Generating candidate regions for RS requires addressing two practical challenges.
First, RS object detection datasets define heterogeneous class taxonomies, so the proposal mechanism must operate flexibly across diverse vocabularies.
Second, using the full class set for every query leads to many irrelevant proposals.

To handle both issues, we adopt a text-conditioned open-vocabulary detection architecture as the backbone of our region proposal module.
Open-vocabulary detection in RS has been explored by standalone systems \cite{openrsd}; our module serves a different role as a component within an MLLM framework, bridging the user's natural language query to the detector's class space.
Specifically, we fine-tune MM-Grounding-DINO \cite{mmgroundingdino} on five RS object detection datasets (xView\cite{xview}, DIOR\cite{dior}, DOTA-v2.0\cite{dota}, FAIR1M\cite{fair1m}, and SODA-A\cite{soda}), constructing a union set of 36 categories that spans diverse RS object types from vehicles and buildings to infrastructure and sports facilities (full class list in Appendix~\ref{sec:app_classes}).
During detector training, only the classes present in each source dataset are provided as text prompts, allowing the model to learn vocabulary-conditioned proposal generation despite the heterogeneous annotations across datasets.
At inference, we compute the semantic similarity between the user query and each class name using a sentence encoder \cite{minilm}, and retain only the top-ranked classes when the similarity distribution is sufficiently peaked, suppressing irrelevant proposals.
The raw outputs are deduplicated, filtered through NMS, and capped at a fixed number of candidates.
Further filtering details are provided in Appendix~\ref{sec:app_rpm}.

\subsection{Region Token Representation}
\label{sec:region_token}

Each candidate $r_k$ is represented by a group of special tokens:
\begin{equation}
\underbrace{\roi{k}}_{\text{proxy token}} \;\; \underbrace{\obj{k,1} \;\; \obj{k,2} \;\; \cdots \;\; \obj{k,M}}_{\text{per-frame visual features}}
\end{equation}
Here, \roi{k} is a learnable proxy token that the LLM can generate in its output to select region $r_k$, and each \obj{k,t} carries the visual features of $r_k$ extracted from image $I_t$.
For bi-temporal tasks ($M{=}2$), each region has two visual slots, enabling the LLM to observe how the same spatial location has changed between time steps.

The visual features are obtained by extracting multi-scale intermediate representations from the vision encoder, applying ROIAlign \cite{roialign} using each candidate's bounding box, and projecting through a learned two-layer MLP to match the LLM's hidden dimension.
The proxy token \roi{k} is augmented with a spatial embedding derived from a geometry descriptor of the bounding box:
\begin{equation}
\begin{split}
\mathbf{g}_k = [&x_1, y_1, x_2, y_2, c_x, c_y, \\
&w, h, a, \log(w/h)]
\end{split}
\end{equation}
\begin{equation}
\begin{split}
\mathbf{e}'_{\roi{k}} &= \mathrm{LN}\Big( \mathrm{WTE}(\roi{k}) \\
&\quad + \alpha_g \cdot \mathrm{GeoMLP}(\mathbf{g}_k) \Big)
\end{split}
\end{equation}
where $(x_1, y_1)$ and $(x_2, y_2)$ are the top-left and bottom-right box coordinates in normalized $[0,1]$ space, $(c_x, c_y)$ is the box center, $w$ and $h$ are the width and height, $a$ is the area, and $\log(w/h)$ encodes the aspect ratio.
$\alpha_g$ is a learnable scalar and LN denotes LayerNorm\cite{layernorm}.
Each visual feature is similarly augmented with a temporal embedding encoding the frame index and total frame count:
\begin{equation}
\mathbf{f}'_{k,t} = \mathrm{LN}\!\left(\mathbf{f}_{k,t} + \alpha_t \cdot (\mathbf{e}^{\mathrm{idx}}_t + \mathbf{e}^{\mathrm{cnt}}_M)\right)
\end{equation}
where $\mathbf{f}_{k,t}$ is the projected visual feature of region $r_k$ from image $I_t$, $\mathbf{e}^{\mathrm{idx}}_t$ is a learnable embedding for the frame index $t$ (indicating which temporal position this image occupies), $\mathbf{e}^{\mathrm{cnt}}_M$ is a learnable embedding for the total number of input frames $M$, and $\alpha_t$ is a learnable scalar analogous to $\alpha_g$.
The spatial cue tells the model where each candidate is located, while the temporal cue lets it distinguish which frame a feature originates from.
The enriched embeddings then replace the corresponding token positions in the LLM input.

\subsection{Region-Aware Language Modeling}
\label{sec:lm}

The LLM receives an input sequence consisting of image patch tokens from the vision encoder, the region token groups for all candidates, and the tokenized text query.
A brief description prefix informs the model of the available region tokens and their temporal ordering.

During generation, the model produces a natural language response that may include \roi{k} tokens to indicate region selections.
Each \roi{k} maps back to its candidate bounding box, yielding the localization result.

Because the model has access to per-frame visual features of the same spatial region, it can attend to both \obj{k,1} (pre-event) and \obj{k,2} (post-event) to assess whether region $r_k$ has changed, enabling visual comparison within the LLM's context.

\subsection{Training}
\label{sec:training}

Our training data spans both multi-temporal and single-image settings.
For multi-temporal tasks, TEOChatlas \cite{teochat} provides change localization (xBD \cite{xbd}, S2Looking \cite{s2looking}), spatial referring expression, change QA, region-level QA, temporal QA (QFabric \cite{qfabric}), and scene classification (fMoW \cite{fmow}).
Additional change detection data comes from LEVIR-MCI \cite{levirmci}, TUE-CD \cite{tuecd}, and other sources \cite{egybcd,s2looking}.
For single-image tasks, GeoChat Instruct \cite{geochat} provides general RS conversation, FIT-RS \cite{fitrs} provides fine-grained understanding, and DIOR-RSVG \cite{diorrsvg} and OPT-RSVG \cite{optrsvg} provide visual grounding.
Including scene-level QA and classification tasks ensures the model builds broad RS domain knowledge, providing the visual understanding foundation on which localization capabilities are built.
Table~\ref{tab:training_data} summarizes dataset usage across training stages.

\begin{figure}[t]
\centering
\includegraphics[width=0.87\columnwidth]{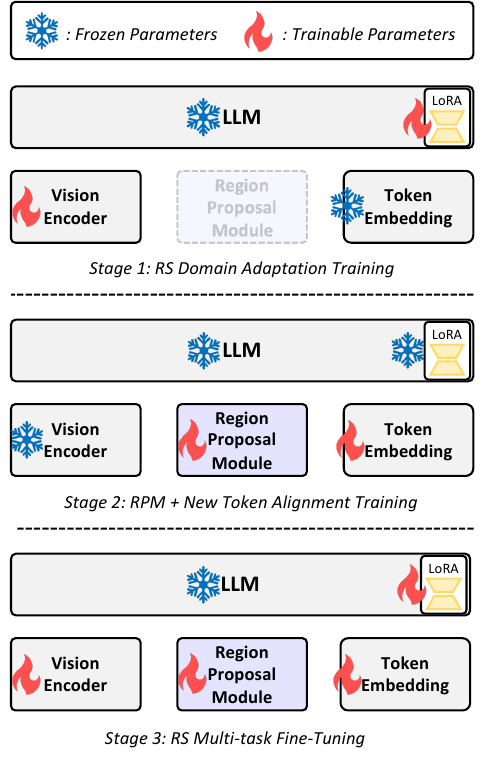}
\caption{Stage-wise training. In Stages 1 and 3, only the visual tokenizer head and the last vision encoder block are trainable within the vision encoder.}
\label{fig:training}
\end{figure}

Training proceeds in three stages (Figure~\ref{fig:training}).
\textbf{Stage~1} (RS domain adaptation) runs for 2 epochs on scene-level RS tasks without any region module, fine-tuning the LLM via LoRA (learning rate $2{\times}10^{-5}$) together with the visual tokenizer head and the last vision encoder block, while the remaining vision encoder layers stay frozen.
\textbf{Stage~2} (RPM and new token alignment training) also runs for 2 epochs and introduces the region proposal module while keeping the LLM frozen to stabilize the newly introduced parameters.
In an initial phase, the region projector and ROI token embeddings are trained (learning rate $1{\times}10^{-4}$) on simple single-image region tasks so the model learns to associate region tokens with visual content.
In a subsequent phase, spatial and temporal embedding components are activated, and training expands to multi-temporal and multi-target settings.
\textbf{Stage~3} (RS multi-task fine-tuning) runs for 2 epochs and jointly trains all components on the full task mixture at a learning rate of $2{\times}10^{-5}$.
The visual embedding components are selectively unfrozen alongside the LLM (via LoRA\cite{lora}), while gradient masking restricts text embedding updates to ROI-specific token rows to preserve pretrained representations.
All LoRA stages use rank 64 and $\alpha{=}128$.
Additional hyperparameters are in Appendix~\ref{sec:app_training}.

\begin{table}[t]
\centering
\small
\begin{tabular}{@{}lccc@{}}
\toprule
\textbf{Dataset Group} & \textbf{1} & \textbf{2} & \textbf{3} \\
\midrule
TEOChatlas \cite{teochat} (scene/QA) & \checkmark & & \checkmark \\
TEOChatlas (region/temporal) & & \checkmark & \checkmark \\
TEOChatlas (localization) & & \checkmark & \checkmark \\
GeoChat Instruct \cite{geochat} & \checkmark & \checkmark & \checkmark \\
RSVG \cite{diorrsvg,optrsvg} & & \checkmark & \checkmark \\
FIT-RS \cite{fitrs} & & & \checkmark \\
\makecell[l]{Additional CD \\ \cite{levirmci,tuecd} \\ \cite{egybcd,s2looking}} & & \checkmark & \checkmark \\
\bottomrule
\end{tabular}
\caption{Dataset usage across training stages (1, 2, 3).}
\label{tab:training_data}
\end{table}

\section{Experiments}
\label{sec:exp}

\subsection{Setup}
\label{sec:setup}

We evaluate on two groups of tasks.
\textbf{Localization tasks} (Table~\ref{tab:loc_results}) form the core evaluation: bi-temporal building localization (LOC) on xBD \cite{xbd}, which asks the model to identify all buildings from temporal image pairs; change detection localization (CDL) on S2Looking \cite{s2looking}, which asks for changed buildings; spatial referring expression (SRE) on both datasets, which grounds a spatially described region; single-image visual grounding on DIOR-RSVG \cite{diorrsvg}, evaluated by box-level Accuracy@0.5; change detection on LEVIR-MCI \cite{levirmci}; and zero-shot change detection on HRCUS-CD \cite{hrcuscd} (not in training data).
All other localization tasks use pixel-level F1.
\textbf{Understanding tasks} (Table~\ref{tab:und_results}) include damage classification on xBD, QA on xBD and S2Looking, temporal QA and temporal referring expression on QFabric \cite{qfabric}, and scene classification on fMoW, evaluated by accuracy except xBD damage classification, which uses F1.
Detailed task descriptions and example prompts are in Appendix~\ref{sec:app_data}.

We compare against four baselines.
\textbf{TEOChat} \cite{teochat} is a temporal RS-MLLM with coordinate generation.
\textbf{EarthDial} \cite{earthdial} is a multi-sensor, multi-temporal RS-MLLM.
\textbf{Qwen3-VL} \cite{qwen3vl} is an off-the-shelf general-purpose MLLM without RS-specific fine-tuning.
\textbf{Ovis2.5-FT} is our base model (Ovis2.5 \cite{ovis}) fine-tuned on the same data with the same recipe, but without the region proposal module, using coordinate generation instead.

\subsection{Main Results}
\label{sec:results}

\paragraph{Localization Tasks.}
Table~\ref{tab:loc_results} presents localization results across seven benchmarks.
The most informative comparison is against Ovis2.5-FT, which shares the same base model, training data, and optimization recipe, differing only in the localization formulation.
On xBD building localization, the gap is striking: our model achieves 69.4\% F1 compared to 32.3\% for Ovis2.5-FT, a 37.1-point improvement from replacing coordinate generation with the complete region-selection interface under the same base model, training data, and optimization recipe.
On S2Looking change detection localization, the gap is 14.1 points (50.5 vs.\ 36.4).
These results confirm that for multi-target temporal localization, where the model must identify many small regions simultaneously, region selection provides a substantial advantage over coordinate generation.

On DIOR-RSVG\cite{diorrsvg} single-image visual grounding, our model reaches 77.3\% accuracy compared to 69.5\% for Ovis2.5-FT, demonstrating that the benefit of region selection extends beyond temporal tasks.
Spatial referring expression (SRE) shows consistent improvement: 40.2\% vs.\ 27.6\% on xBD and 49.6\% vs.\ 36.1\% on S2Looking, though absolute scores remain moderate for all models, reflecting the inherent difficulty of grounding complex spatial descriptions in RS imagery.
HRCUS-CD, evaluated in a zero-shot setting (not in training data), yields 55.0\% F1, and LEVIR-MCI reaches 58.8\%, both surpassing the coordinate-generation baseline.
Comparisons with two recent open-weight generalist MLLMs, Qwen3.5-9B~\cite{qwen3.5} and InternVL3.5-8B~\cite{internvl3.5}, are reported in Appendix~\ref{app:generalist}, where our model leads on all five evaluated localization benchmarks.

Compared to existing RS-MLLMs, our model outperforms TEOChat \cite{teochat} on all localization benchmarks despite TEOChat having been specifically designed for temporal RS tasks.
EarthDial \cite{earthdial} and Qwen3-VL \cite{qwen3vl} show substantially lower scores, particularly on datasets they were not trained on (marked with $^*$ in the table).

\begin{figure*}[!t]
\centering
\includegraphics[width=\textwidth]{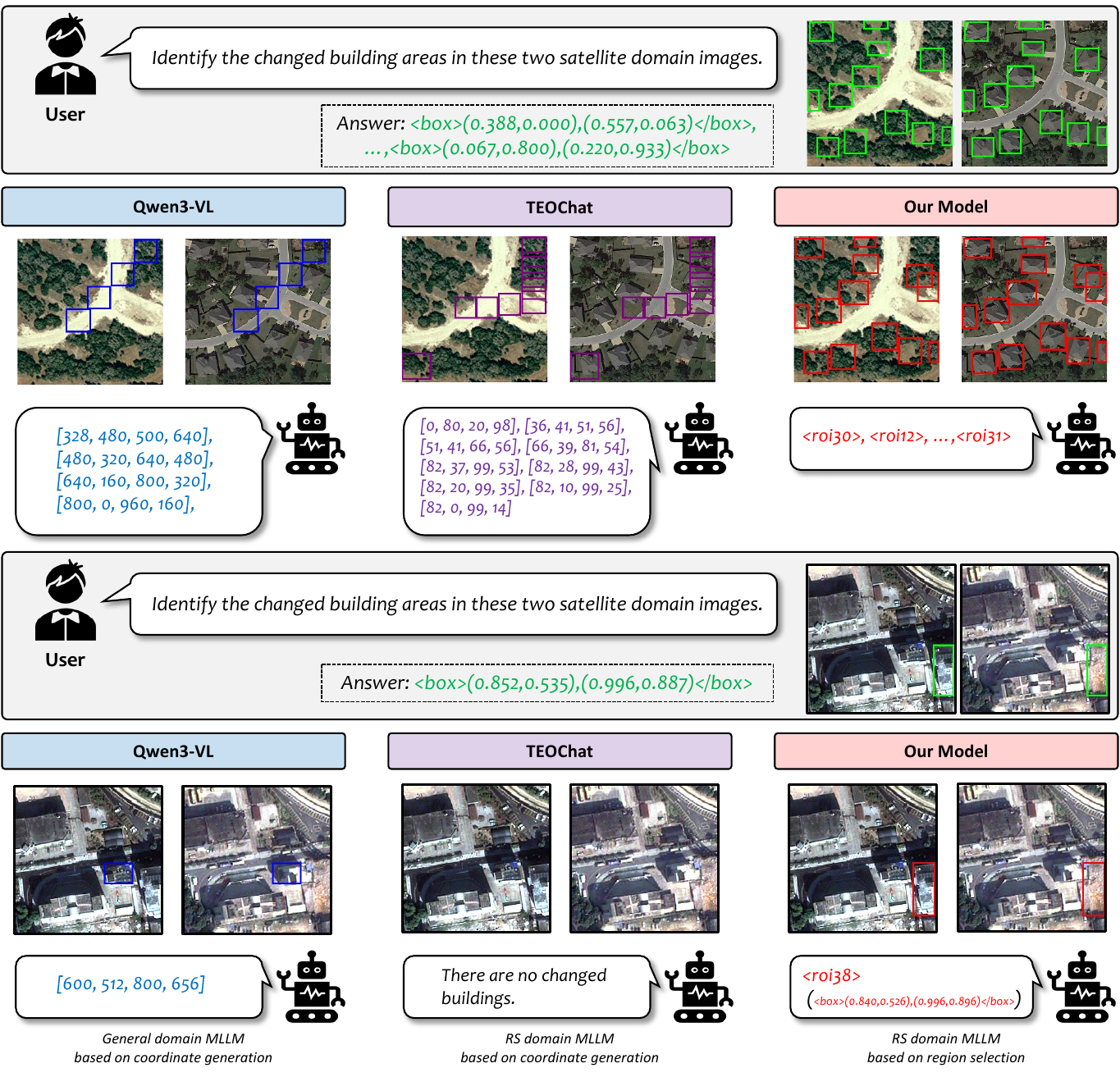}
\caption{Qualitative comparison of change detection localization. \textbf{Top (LEVIR-MCI)}\cite{levirmci}: Qwen3-VL\cite{qwen3vl} (coordinates normalized to 0--1000) outputs a degenerate arithmetic pattern. TEOChat\cite{teochat} (0--100) produces many small boxes. Our model identifies changed regions through region token selection. \textbf{Bottom (HRCUS-CD}\cite{hrcuscd}\textbf{, zero-shot):} Qwen3-VL generates mislocated coordinates. TEOChat fails to detect change. Our model selects \roi{38}. Green boxes (top-right) show ground truth (0--1 normalized).}
\label{fig:qualitative}
\end{figure*}

Figure~\ref{fig:qualitative} illustrates two representative examples.
On a LEVIR-MCI \cite{levirmci} scene with many changed buildings (top), Qwen3-VL \cite{qwen3vl} produces coordinates in a degenerate arithmetic pattern, while TEOChat \cite{teochat} generates many small boxes with limited coverage.
On an HRCUS-CD \cite{hrcuscd} scene (bottom, zero-shot), Qwen3-VL generates mislocated coordinates and TEOChat fails to detect any change.
In both cases, our model correctly identifies the relevant regions through region token selection.

\begin{table*}[t]
\centering
\resizebox{\textwidth}{!}{
\begin{tabular}{@{}llccccccc@{}}
\toprule
& & \textbf{xBD} & \textbf{S2Looking} & \textbf{xBD} & \textbf{S2Looking} & \textbf{DIOR-RSVG} & \textbf{HRCUS-CD} & \textbf{LEVIR-MCI} \\
\textbf{Model} & \textbf{Size} & LOC (F1) & CDL (F1) & SRE (F1) & SRE (F1) & Acc@0.5 & F1 & F1 \\
\midrule
Qwen3-VL \cite{qwen3vl} & 8B & 17.3$^*$ & 11.8$^*$ & 6.7$^*$ & 7.9$^*$ & 53.8$^*$ & 21.7$^*$ & 13.3$^*$ \\
TEOChat \cite{teochat} & 7B & 38.9 & 34.5 & 25.1 & 32.9 & 27.6 & 33.3$^*$ & 26.4$^*$ \\
EarthDial \cite{earthdial} & 4B & 24.1 & 2.6$^*$ & 8.7 & 12.2$^*$ & 39.6 & 4.7$^*$ & 12.6$^*$ \\
Ovis2.5-FT & 9B & 32.3 & 36.4 & 27.6 & 36.1 & 69.5 & 54.4$^\dag$ & 56.4 \\
\textbf{Ours} & 9B & \textbf{69.4} & \textbf{50.5} & \textbf{40.2} & \textbf{49.6} & \textbf{77.3} & \textbf{55.0}$^\dag$ & \textbf{58.8} \\
\bottomrule
\end{tabular}%
}
\caption{Localization results (\%). LOC: bi-temporal building localization; CDL: change detection localization; SRE: spatial referring expression. Our model achieves the best performance on all benchmarks. $^*$Zero-shot (not trained on this dataset). $^\dag$HRCUS-CD not in training data.}
\label{tab:loc_results}
\end{table*}

\paragraph{Understanding Tasks.}
Table~\ref{tab:und_results} presents understanding results.
Our model performs comparably to TEOChat \cite{teochat} on scene-level QA (xBD QA: 85.9 vs.\ 89.9; S2Looking QA: 74.0 vs.\ 73.4), confirming that the region selection interface does not degrade general understanding capabilities.
On QFabric\cite{qfabric} temporal tasks, where region-level visual features directly aid reasoning about specific spatial regions across time, our model achieves the best results (TRE: 78.0\%, RTQA: 78.6\%), outperforming both TEOChat and Ovis2.5-FT.
Scene classification on fMoW (73.5\%) is slightly below TEOChat (75.1\%) but substantially above Qwen3-VL\cite{qwen3vl} (37.3\%) and EarthDial\cite{earthdial} (37.2\%).
Notably, the improvement over Ovis2.5-FT on temporal tasks (QFabric TRE: 78.0 vs.\ 73.5; RTQA: 78.6 vs.\ 75.6) suggests that region-level features benefit not only localization but also region-conditioned understanding.

\begin{table}[t]
\centering
\resizebox{\columnwidth}{!}{%
\begin{tabular}{@{}lcccccc@{}}
\toprule
\textbf{Model} & \rotatebox{55}{\textbf{fMoW}} & \rotatebox{55}{\textbf{xBD CDC}} & \rotatebox{55}{\textbf{xBD QA}} & \rotatebox{55}{\textbf{S2L QA}} & \rotatebox{55}{\textbf{QF TRE}} & \rotatebox{55}{\textbf{QF RTQA}} \\
\midrule
Qwen3-VL   & 37.3 & 34.6 & 44.7 & 41.3 & 22.1 & 61.6 \\
TEOChat    & \textbf{75.1} & 50.0 & \textbf{89.9} & 73.4 & 74.9 & 71.7 \\
EarthDial  & 37.2 & ~0.6 & 37.4 & 48.8 & ~0.0 & 53.7 \\
Ovis2.5-FT & 71.2 & 45.2 & 85.8 & 74.0 & 73.5 & 75.6 \\
\textbf{Ours}       & 73.5 & \textbf{51.2} & 85.9 & \textbf{74.0} & \textbf{78.0} & \textbf{78.6} \\
\bottomrule
\end{tabular}%
}
\caption{Understanding results (\%). xBD CDC is F1 and the remaining tasks are accuracy. CDC: change damage classification; S2L: S2Looking; QF: QFabric; TRE: temporal referring expression; RTQA: region temporal QA. Our model shows no region interface degradation and leads temporal region tasks.}
\label{tab:und_results}
\end{table}

\subsection{Analysis}
\label{sec:analysis}

\paragraph{Ablation Study.}
Table~\ref{tab:ablation} isolates the contribution of each component.

\emph{Full model vs.\ No ROI (Ovis2.5-FT).}
Reverting to coordinate generation with the same architecture and data: S2Looking\cite{s2looking} CDL drops from 50.5 to 36.4 and DIOR-RSVG\cite{diorrsvg} from 77.3 to 69.5.

\emph{Full model vs.\ Text-coord ROI.}
Replacing visual features with textual coordinates while retaining selection.
Change detection drops moderately (S2Looking CDL: 50.5$\to$47.5), while DIOR-RSVG remains strong (74.7) since text coordinates suffice for spatial information in single-image tasks.

\emph{Full model vs.\ Visual feature only (w/o spatial and temporal cues).}
DIOR-RSVG drops dramatically (77.3$\to$32.3): queries describe objects by spatial attributes (e.g., ``the small ship on the right''), and without the geometry embedding the model can recognize visual content but cannot determine a candidate's location or relative size, making it unable to match the spatial description.
Separate ablations show that this drop is driven mainly by the spatial cue, while the temporal cue provides a smaller but consistent gain, with DIOR-RSVG dropping from 77.3 to 33.5 without the spatial cue. The full six-configuration grid is reported in Appendix~\ref{app:full_ablation}.

\begin{table}[t]
\centering
\small
\setlength{\tabcolsep}{2.5pt}
\begin{tabular}{@{}lccccc@{}}
\toprule
& \textbf{S2L} & \textbf{S2L} & \textbf{RSVG} & \textbf{LEVIR} & \textbf{HRCUS} \\
\textbf{Config.} & CDL & SRE & Acc & F1 & F1 \\
\midrule
Full model          & \textbf{50.5} & \textbf{49.6} & \textbf{77.3} & \textbf{58.8} & \textbf{55.0} \\
w/o spat.+temp.     & 39.0 & 39.3 & 32.3 & 56.6 & 41.6 \\
Text-coord ROI      & 47.5 & 47.9 & 74.7 & 55.7 & 50.1 \\
No ROI (coord gen)  & 36.4 & 36.1 & 69.5 & 56.4 & 54.4 \\
\bottomrule
\end{tabular}
\caption{Ablation study (\%). S2L: S2Looking; CDL: change detection localization; SRE: spatial referring expression; RSVG: DIOR-RSVG (Acc@0.5).}
\label{tab:ablation}
\end{table}

\paragraph{Oracle and Proposer Recall Analysis.}
A distinctive advantage of the region selection framework is that it cleanly decomposes end-to-end performance into three interpretable levels: \emph{proposer recall} (AR@100, measuring what fraction of ground-truth regions appear among the candidates), \emph{oracle F1} (model performance when ground-truth boxes are added to the candidate set, guaranteeing perfect recall), and \emph{standard F1} (actual end-to-end performance).
Table~\ref{tab:oracle} presents this decomposition.

The proposer, trained solely on five RS object detection datasets that contain no change detection annotations, generalizes remarkably well in a zero-shot manner to temporal change benchmarks: AR@100 reaches 74.3\% on xBD\cite{xbd}, 70.3\% on S2Looking\cite{s2looking}, 78.7\% on HRCUS-CD\cite{hrcuscd}, and 87.1\% on LEVIR-MCI\cite{levirmci}.
This indicates that the unified 36-class vocabulary learned from heterogeneous RS detection datasets transfers effectively to building-centric localization tasks.

The decomposition reveals that the performance bottleneck varies by dataset.
On xBD, the proposer recall is 74.3\% and the standard F1 reaches 69.4\%, which is close to the oracle F1 of 76.3\% (gap: 6.9 points).
This means the LLM selector is effective at utilizing the available candidates, and the primary ceiling comes from proposals that the detector misses.
On LEVIR-MCI, in contrast, the proposer achieves the highest recall (87.1\%), yet the standard F1 is only 58.8\% compared to 70.0\% oracle (gap: 11.2 points).
Here the selector, not the proposer, is the dominant bottleneck: the detector finds most targets, but the LLM fails to select all of them correctly.
On S2Looking, both the proposer recall (70.3\%) and the standard-to-oracle gap (9.8 points) indicate room for improvement on both sides.

For DIOR-RSVG\cite{diorrsvg}, where the proposer was trained on DIOR\cite{dior} data (marked $^\ddag$), recall is naturally higher (83.0\%) and the oracle gap is small (77.3 vs.\ 80.7), suggesting that the framework is operating near its ceiling for this task.

This analysis provides actionable guidance: for datasets where the oracle gap is large (e.g., LEVIR-MCI), improving the LLM's selection ability through better training data or objectives is the priority; for datasets where proposer recall is the ceiling (e.g., S2Looking), strengthening the detector or expanding its training data is more impactful.
This kind of decomposition is not available in coordinate-generation models, where the quality of localization and the model's spatial reasoning are entangled and cannot be separately diagnosed.

\begin{table}[t]
\centering
\small
\setlength{\tabcolsep}{3pt}
\begin{tabular}{@{}lccc@{}}
\toprule
\textbf{Task} & \textbf{AR@100} & \textbf{Standard} & \textbf{Oracle} \\
\midrule
xBD LOC (F1)        & 74.3 & 69.4 & 76.3 \\
S2Looking CDL (F1)   & 70.3 & 50.5 & 60.3 \\
xBD SRE (F1)        & 74.3 & 40.2 & 50.2 \\
S2Looking SRE (F1)   & 70.3 & 49.6 & 58.3 \\
DIOR-RSVG (Acc)$^\ddag$ & 83.0 & 77.3 & 80.7 \\
HRCUS-CD (F1)       & 78.7 & 55.0 & 61.1 \\
LEVIR-MCI (F1)      & 87.1 & 58.8 & 70.0 \\
\bottomrule
\end{tabular}
\caption{Three-level performance decomposition (\%). AR@100: proposer recall (fraction of GT regions found among candidates at IoU${\geq}$0.5). Standard: end-to-end model performance. Oracle: performance when GT boxes are added to the candidate set, guaranteeing perfect proposer recall. $^\ddag$Proposer trained on DIOR; all others zero-shot.}
\label{tab:oracle}
\end{table}

\paragraph{Temporal Comparison Diagnostics.}
To verify that localization reflects cross-frame comparison rather than appearance-only selection, we replace each evaluation pair with an identical pair by duplicating the post-event image into both temporal slots. Since post-event objects remain visible, an appearance-only model would still emit regions. As shown in Table~\ref{tab:identical}, both the region-token emission rate and the number of predicted regions collapse to near zero on identical pairs. Reversing the actual image order changes 41.2\% of responses on QFabric, whereas reversing only the frame-index embeddings changes TRE by a single point (78.0 to 77.0). This indicates that temporal behavior is driven by ordered per-frame features, and the frame-index embedding acts as a lightweight identifier.

\begin{table}[t]
\centering\small
\setlength{\tabcolsep}{4.5pt}
\begin{tabular}{lcccc}
\toprule
& \multicolumn{2}{c}{Emission rate (\%)} & \multicolumn{2}{c}{Avg.\ \#regions} \\
\cmidrule(lr){2-3}\cmidrule(lr){4-5}
 & Normal & Identical & Normal & Identical \\
\midrule
S2Looking & 74.6 & 2.8 & 1.35 & 0.03 \\
LEVIR-MCI & 39.1 & 0.0 & 3.06 & 0.00 \\
\bottomrule
\end{tabular}
\caption{Identical-pair control. Each test pair is replaced with a no-change pair by duplicating the post-event image into both temporal slots. Region-token emission collapses to near zero, with no emission on any of the 1{,}929 LEVIR-MCI pairs.}
\label{tab:identical}
\end{table}

\paragraph{Target-Density Analysis.}
Coordinate generation degrades sharply as targets multiply. On LEVIR-MCI, TEOChat falls from 47.6 F1 for a single target to 26.2 for more than 20 targets, while our model remains at 58.1 in the densest bin (Table~\ref{tab:density}). Degenerate arithmetic coordinate patterns appear in 19.0\% to 28.8\% of Qwen3-VL outputs and in 9.7\% of Ovis2.5-FT outputs. Such failure modes cannot arise under region selection, which produces discrete token choices instead of coordinate sequences.

\begin{table}[t]
\centering\small
\begin{tabular}{lccccc}
\toprule
GT targets & 1 & 2--5 & 6--10 & 11--20 & $>$20 \\
\midrule
TEOChat & 47.6 & 38.9 & 30.6 & 26.9 & 26.2 \\
\textbf{Ours}    & \textbf{51.3} & \textbf{60.6} & \textbf{59.4} & \textbf{59.6} & \textbf{58.1} \\
\bottomrule
\end{tabular}
\caption{Change localization F1 on LEVIR-MCI, grouped by the number of ground-truth changed regions per sample. TEOChat degrades as targets multiply, while our model remains stable.}
\label{tab:density}
\end{table}

\section{Conclusion}
\label{sec:conclusion}

We have presented an RS-specific formulation of the region selection paradigm for remote sensing localization and multi-temporal change localization.
By encoding each candidate region with per-frame visual features, spatial geometry, and temporal cues, the model performs localization through discrete token selection rather than coordinate generation.
In controlled experiments where the only difference is the localization formulation, region selection yields substantial improvements on change detection localization and visual grounding tasks.
Understanding tasks remain competitive, confirming that the region interface does not degrade general capabilities.
Oracle analysis provides a clean decomposition of proposer and selector contributions, a diagnostic unique to this framework that offers concrete guidance for future improvement.

\section*{Limitations}

\paragraph{Proposer Recall Ceiling.}
The model can only select from candidates generated by the region proposal module; regions not proposed cannot be recovered.
Our oracle analysis quantifies this ceiling and shows that it is a significant factor, particularly on S2Looking CDL (9.8-point gap) and LEVIR-MCI (11.2-point gap).
Improving proposer coverage through stronger detectors, iterative proposals, or expanded training data is a natural direction for future work.

\paragraph{Training-Evaluation Metric Gap.}
The model is trained with token-level cross-entropy loss but evaluated with set-level F1 at a fixed IoU threshold.
This mismatch means that predicting the correct set of regions in a different order incurs a training loss despite being equally valid.
Exploring set-level objectives or reinforcement learning with task-specific rewards could help bridge this gap.

\paragraph{Future Directions.}
The current framework operates at the axis-aligned bounding box level.
Extending it to oriented bounding boxes or pixel-level masks would broaden its applicability to a wider range of RS tasks, provided sufficient training data is available.
On the temporal side, our diagnostics support region-level cross-frame comparison between aligned candidate features rather than broader temporal reasoning over long sequences, and extending the framework to longer sequences and richer temporal relations is another natural direction.

\section*{Acknowledgments}
This work was supported in part by the Institute of Information and Communications Technology Planning and Evaluation (IITP) grant funded by the Korean Government (Ministry of Science and ICT) under Grant RS-2022-II220124, and in part by the IITP grant funded by the Korean Government (Ministry of Science and ICT) under Grant RS-2022-II220984.

\bibliography{custom}

@inproceedings{geochat,
  title={Geochat: Grounded large vision-language model for remote sensing},
  author={Kuckreja, Kartik and Danish, Muhammad Sohail and Naseer, Muzammal and Das, Abhijit and Khan, Salman and Khan, Fahad Shahbaz},
  booktitle={Proceedings of the IEEE/CVF conference on computer vision and pattern recognition},
  pages={27831--27840},
  year={2024}
}

@inproceedings{teochat,
  title={TEOChat: A Large Vision-Language Assistant for Temporal Earth Observation Data},
  author={Irvin, Jeremy Andrew and Liu, Emily Ruoyu and Chen, Joyce Chuyi and Dormoy, Ines and Kim, Jinyoung and Khanna, Samar and Zheng, Zhuo and Ermon, Stefano},
  booktitle={International Conference on Learning Representations},
  year={2025}
}

@inproceedings{earthdial,
  title     = {EarthDial: Turning Multi-sensory Earth Observations to Interactive Dialogues},
  author    = {Soni, Sagar and Dudhane, Akshay and Debary, Hiyam and Fiaz, Mustansar and Munir, Muhammad Akhtar and Danish, Muhammad Sohail and Fraccaro, Paolo and Watson, Campbell D and Klein, Levente J and Khan, Fahad Shahbaz and Khan, Salman},
  booktitle = {Proceedings of the IEEE/CVF Conference on Computer Vision and Pattern Recognition (CVPR)},
  year      = {2025}
}

@inproceedings{lhrsbot,
  title={Lhrs-bot: Empowering remote sensing with vgi-enhanced large multimodal language model},
  author={Muhtar, Dilxat and Li, Zhenshi and Gu, Feng and Zhang, Xueliang and Xiao, Pengfeng},
  booktitle={European Conference on Computer Vision},
  pages={440--457},
  year={2024},
  organization={Springer}
}

@article{earthgpt,
  title={EarthGPT: A universal multimodal large language model for multisensor image comprehension in remote sensing domain},
  author={Zhang, Wei and Cai, Miaoxin and Zhang, Tong and Zhuang, Yin and Mao, Xuerui},
  journal={IEEE Transactions on Geoscience and Remote Sensing},
  volume={62},
  pages={1--20},
  year={2024},
  publisher={IEEE}
}

@article{skyeyegpt,
  title={Skyeyegpt: Unifying remote sensing vision-language tasks via instruction tuning with large language model},
  author={Zhan, Yang and Xiong, Zhitong and Yuan, Yuan},
  journal={ISPRS Journal of Photogrammetry and Remote Sensing},
  volume={221},
  pages={64--77},
  year={2025},
  publisher={Elsevier}
}

@article{geoground,
  title={GeoGround: A unified large vision-language model for remote sensing visual grounding},
  author={Zhou, Yue and Lan, Mengcheng and Li, Xiang and Feng, Litong and Ke, Yiping and Jiang, Xue and Li, Qingyun and Yang, Xue and Zhang, Wayne},
  journal={arXiv preprint arXiv:2411.11904},
  year={2024}
}

@article{geopixel,
  title={Geopixel: Pixel grounding large multimodal model in remote sensing},
  author={Shabbir, Akashah and Zumri, Mohammed and Bennamoun, Mohammed and Khan, Fahad S and Khan, Salman},
  journal={arXiv preprint arXiv:2501.13925},
  year={2025}
}

@article{terrascope,
  title={TerraScope: Pixel-Grounded Visual Reasoning for Earth Observation},
  author={Shu, Yan and Ren, Bin and Xiong, Zhitong and Zhu, Xiao Xiang and Demir, Beg{\"u}m and Sebe, Nicu and Rota, Paolo},
  journal={arXiv preprint arXiv:2603.19039},
  year={2026}
}

@inproceedings{groma,
  title={Groma: Localized visual tokenization for grounding multimodal large language models},
  author={Ma, Chuofan and Jiang, Yi and Wu, Jiannan and Yuan, Zehuan and Qi, Xiaojuan},
  booktitle={European Conference on Computer Vision},
  pages={417--435},
  year={2024},
  organization={Springer}
}

@article{chatrex,
  title={Chatrex: Taming multimodal llm for joint perception and understanding},
  author={Jiang, Qing and Luo, Gen and Yang, Yuqin and Xiong, Yuda and Chen, Yihao and Zeng, Zhaoyang and Ren, Tianhe and Zhang, Lei},
  journal={arXiv preprint arXiv:2411.18363},
  year={2024}
}

@inproceedings{gpt4roi,
  title={Gpt4roi: Instruction tuning large language model on region-of-interest},
  author={Zhang, Shilong and Sun, Peize and Chen, Shoufa and Xiao, Min and Shao, Wenqi and Zhang, Wenwei and Liu, Yu and Chen, Kai and Luo, Ping},
  booktitle={European Conference on Computer Vision},
  pages={52--70},
  year={2025},
  organization={Springer}
}

@inproceedings{osprey,
  title={Osprey: Pixel understanding with visual instruction tuning},
  author={Yuan, Yuqian and Li, Wentong and Liu, Jian and Tang, Dongqi and Luo, Xinjie and Qin, Chi and Zhang, Lei and Zhu, Jianke},
  booktitle={Proceedings of the IEEE/CVF Conference on Computer Vision and Pattern Recognition},
  pages={28202--28211},
  year={2024}
}

@inproceedings{changechat,
  title={Changechat: An interactive model for remote sensing change analysis via multimodal instruction tuning},
  author={Deng, Pei and Zhou, Wenqian and Wu, Hanlin},
  booktitle={ICASSP 2025-2025 IEEE International Conference on Acoustics, Speech and Signal Processing (ICASSP)},
  pages={1--5},
  year={2025},
  organization={IEEE}
}

@inproceedings{cdchat,
  title={Cdchat: A large multimodal model for remote sensing change description},
  author={Noman, Mubashir and Ahsan, Noor and Naseer, Muzammal and Cholakkal, Hisham and Anwer, Rao Muhammad and Khan, Salman and Khan, Fahad Shahbaz},
  booktitle={IGARSS 2025-2025 IEEE International Geoscience and Remote Sensing Symposium},
  pages={7033--7037},
  year={2025},
  organization={IEEE}
}

@article{shikra,
  title={Shikra: Unleashing multimodal llm's referential dialogue magic},
  author={Chen, Keqin and Zhang, Zhao and Zeng, Weili and Zhang, Richong and Zhu, Feng and Zhao, Rui},
  journal={arXiv preprint arXiv:2306.15195},
  year={2023}
}

@article{qwen2,
  title   = {Qwen2-VL: Enhancing Vision-Language Model's Perception of the World at Any Resolution},
  author  = {Wang, Peng and Bai, Shuai and Tan, Sinan and Wang, Shijie and Fan, Zhihao and Bai, Jinze and Chen, Keqin and Liu, Xuejing and Wang, Jialin and Ge, Wenbin and Fan, Yang and Dang, Kai and Du, Mengfei and Ren, Xuancheng and Men, Rui and Liu, Dayiheng and Zhou, Chang and Zhou, Jingren and Lin, Junyang},
  journal = {arXiv preprint arXiv:2409.12191},
  year    = {2024}
}

@inproceedings{lisa,
  title={Lisa: Reasoning segmentation via large language model},
  author={Lai, Xin and Tian, Zhuotao and Chen, Yukang and Li, Yanwei and Yuan, Yuhui and Liu, Shu and Jia, Jiaya},
  booktitle={Proceedings of the IEEE/CVF conference on computer vision and pattern recognition},
  pages={9579--9589},
  year={2024}
}

@article{diorrsvg,
  title={Rsvg: Exploring data and models for visual grounding on remote sensing data},
  author={Zhan, Yang and Xiong, Zhitong and Yuan, Yuan},
  journal={IEEE transactions on geoscience and remote sensing},
  volume={61},
  pages={1--13},
  year={2023},
  publisher={IEEE}
}

@article{ovis,
  title   = {Ovis2.5 Technical Report},
  author  = {Lu, Shiyin and Li, Yang and Xia, Yu and Hu, Yuwei and Zhao, Shanshan and Ma, Yanqing and Wei, Zhichao and Li, Yinglun and Duan, Lunhao and Zhao, Jianshan and Han, Yuxuan and Li, Haijun and Chen, Wanying and Tang, Junke and Hou, Chengkun and Du, Zhixing and Zhou, Tianli and Zhang, Wenjie and Ding, Huping and Li, Jiahe and Li, Wen and Hu, Gui and Gu, Yiliang and Yang, Siran and Wang, Jiamang and Sun, Hailong and Wang, Yibo and Sun, Hui and Huang, Jinlong and He, Yuping and Shi, Shengze and Zhang, Weihong and Zheng, Guodong and Jiang, Junpeng and Gao, Sensen and Wu, Yi-Feng and Chen, Sijia and Chen, Yuhui and Chen, Qing-Guo and Xu, Zhao and Luo, Weihua and Zhang, Kaifu},
  journal = {arXiv preprint arXiv:2508.11737},
  year    = {2025}
}

@article{siglip2,
  title   = {SigLIP 2: Multilingual Vision-Language Encoders with Improved Semantic Understanding, Localization, and Dense Features},
  author  = {Tschannen, Michael and Gritsenko, Alexey and Wang, Xiao and Naeem, Muhammad Ferjad and Alabdulmohsin, Ibrahim and Parthasarathy, Nikhil and Evans, Talfan and Beyer, Lucas and Xia, Ye and Mustafa, Basil and H{\'e}naff, Olivier and Harmsen, Jeremiah and Steiner, Andreas and Zhai, Xiaohua},
  journal = {arXiv preprint arXiv:2502.14786},
  year    = {2025}
}

@inproceedings{navit,
  title     = {Patch n' Pack: NaViT, a Vision Transformer for any Aspect Ratio and Resolution},
  author    = {Dehghani, Mostafa and Mustafa, Basil and Djolonga, Josip and Heek, Jonathan and Minderer, Matthias and Caron, Mathilde and Steiner, Andreas and Puigcerver, Joan and Geirhos, Robert and Alabdulmohsin, Ibrahim and Oliver, Avital and Padlewski, Piotr and Gritsenko, Alexey and Lu{\v{c}}i{\'c}, Mario and Houlsby, Neil},
  booktitle = {Advances in Neural Information Processing Systems (NeurIPS)},
  volume    = {36},
  year      = {2023}
}

@article{qwen3,
  title   = {Qwen3 Technical Report},
  author  = {Yang, An and Li, Anfeng and Yang, Baosong and Zhang, Beichen and Hui, Binyuan and Zheng, Bo and Yu, Bowen and Gao, Chang and Huang, Chengen and Lv, Chenxu and Zheng, Chujie and Liu, Dayiheng and Zhou, Fan and Huang, Fei and Hu, Feng and Ge, Hao and Wei, Haoran and Lin, Huan and Tang, Jialong and Yang, Jian and Tu, Jianhong and Zhang, Jianwei and Yang, Jianxin and Yang, Jiaxi and Zhou, Jing and Zhou, Jingren and Lin, Junyang and Dang, Kai and Bao, Keqin and Yang, Kexin and Yu, Le and Deng, Lianghao and Li, Mei and Xue, Mingfeng and Li, Mingze and Zhang, Pei and Wang, Peng and Zhu, Qin and Men, Rui and Gao, Ruize and Liu, Shixuan and Luo, Shuang and Li, Tianhao and Tang, Tianyi and Yin, Wenbiao and Ren, Xingzhang and Wang, Xinyu and Zhang, Xinyu and Ren, Xuancheng and Fan, Yang and Su, Yang and Zhang, Yichang and Zhang, Yinger and Wan, Yu and Liu, Yuqiong and Wang, Zekun and Cui, Zeyu and Zhang, Zhenru and Zhou, Zhipeng and Qiu, Zihan},
  journal = {arXiv preprint arXiv:2505.09388},
  year    = {2025}
}

@article{mmgroundingdino,
  title={An open and comprehensive pipeline for unified object grounding and detection},
  author={Zhao, Xiangyu and Chen, Yicheng and Xu, Shilin and Li, Xiangtai and Wang, Xinjiang and Li, Yining and Huang, Haian},
  journal={arXiv preprint arXiv:2401.02361},
  year={2024}
}

@article{minilm,
  title={Minilm: Deep self-attention distillation for task-agnostic compression of pre-trained transformers},
  author={Wang, Wenhui and Wei, Furu and Dong, Li and Bao, Hangbo and Yang, Nan and Zhou, Ming},
  journal={Advances in neural information processing systems},
  volume={33},
  pages={5776--5788},
  year={2020}
}

@inproceedings{roialign,
  title={Mask r-cnn},
  author={He, Kaiming and Gkioxari, Georgia and Doll{\'a}r, Piotr and Girshick, Ross},
  booktitle={Proceedings of the IEEE international conference on computer vision},
  pages={2961--2969},
  year={2017}
}

@article{lora,
  title={Lora: Low-rank adaptation of large language models},
  author={Hu, Edward J and Shen, Yelong and Wallis, Phillip and Allen-Zhu, Zeyuan and Li, Yuanzhi and Wang, Shean and Wang, Lu and Chen, Weizhu},
  journal={arXiv preprint arXiv:2106.09685},
  year={2021}
}

@article{fitrs,
  title   = {SkySenseGPT: A Fine-Grained Instruction Tuning Dataset and Model for Remote Sensing Vision-Language Understanding},
  author  = {Luo, Junwei and Pang, Zhen and Zhang, Yongjun and Wang, Tingzhu and Wang, Linlin and Dang, Bo and Lao, Jiangwei and Wang, Jian and Chen, Jingdong and Tan, Yihua and Li, Yansheng},
  journal = {arXiv preprint arXiv:2406.10100},
  year    = {2024}
}

@article{xview,
  title={xview: Objects in context in overhead imagery},
  author={Lam, Darius and Kuzma, Richard and McGee, Kevin and Dooley, Samuel and Laielli, Michael and Klaric, Matthew and Bulatov, Yaroslav and McCord, Brendan},
  journal={arXiv preprint arXiv:1802.07856},
  year={2018}
}

@article{dior,
  title={Object detection in optical remote sensing images: A survey and a new benchmark},
  author={Li, Ke and Wan, Gang and Cheng, Gong and Meng, Liqiu and Han, Junwei},
  journal={ISPRS journal of photogrammetry and remote sensing},
  volume={159},
  pages={296--307},
  year={2020},
  publisher={Elsevier}
}

@ARTICLE{dota,
  author={Ding, Jian and Xue, Nan and Xia, Gui-Song and Bai, Xiang and Yang, Wen and Yang, Michael Ying and Belongie, Serge and Luo, Jiebo and Datcu, Mihai and Pelillo, Marcello and Zhang, Liangpei},
  journal={IEEE Transactions on Pattern Analysis and Machine Intelligence}, 
  title={Object Detection in Aerial Images: A Large-Scale Benchmark and Challenges}, 
  year={2022},
  volume={44},
  number={11},
  pages={7778-7796},
  doi={10.1109/TPAMI.2021.3117983}}

@article{fair1m,
  title   = {FAIR1M: A benchmark dataset for fine-grained object recognition in high-resolution remote sensing imagery},
  author  = {Sun, Xian and Wang, Peijin and Yan, Zhiyuan and Xu, Feng and Wang, Ruiping and Diao, Wenhui and Chen, Jin and Li, Jihao and Feng, Yingchao and Xu, Tao and Weinmann, Martin and Hinz, Stefan and Wang, Cheng and Fu, Kun},
  journal = {ISPRS Journal of Photogrammetry and Remote Sensing},
  volume  = {184},
  pages   = {116--130},
  year    = {2022},
  publisher = {Elsevier}
}

@article{soda,
  title={Towards large-scale small object detection: Survey and benchmarks},
  author={Cheng, Gong and Yuan, Xiang and Yao, Xiwen and Yan, Kebing and Zeng, Qinghua and Xie, Xingxing and Han, Junwei},
  journal={IEEE transactions on pattern analysis and machine intelligence},
  volume={45},
  number={11},
  pages={13467--13488},
  year={2023},
  publisher={IEEE}
}

@inproceedings{sentencebert,
  title={Sentence-bert: Sentence embeddings using siamese bert-networks},
  author={Reimers, Nils and Gurevych, Iryna},
  booktitle={Proceedings of the 2019 conference on empirical methods in natural language processing and the 9th international joint conference on natural language processing (EMNLP-IJCNLP)},
  pages={3982--3992},
  year={2019}
}

@article{xbd,
  title={xbd: A dataset for assessing building damage from satellite imagery},
  author={Gupta, Ritwik and Hosfelt, Richard and Sajeev, Sandra and Patel, Nirav and Goodman, Bryce and Doshi, Jigar and Heim, Eric and Choset, Howie and Gaston, Matthew},
  journal={arXiv preprint arXiv:1911.09296},
  year={2019}
}

@article{s2looking,
  title={S2Looking: A satellite side-looking dataset for building change detection},
  author={Shen, Li and Lu, Yao and Chen, Hao and Wei, Hao and Xie, Donghai and Yue, Jiabao and Chen, Rui and Lv, Shouye and Jiang, Bitao},
  journal={Remote Sensing},
  volume={13},
  number={24},
  pages={5094},
  year={2021},
  publisher={MDPI}
}

@inproceedings{fmow,
  title={Functional map of the world},
  author={Christie, Gordon and Fendley, Neil and Wilson, James and Mukherjee, Ryan},
  booktitle={Proceedings of the IEEE Conference on Computer Vision and Pattern Recognition},
  pages={6172--6180},
  year={2018}
}

@INPROCEEDINGS{qfabric,
  author={Verma, Sagar and Panigrahi, Akash and Gupta, Siddharth},
  booktitle={2021 IEEE/CVF Conference on Computer Vision and Pattern Recognition Workshops (CVPRW)}, 
  title={QFabric: Multi-Task Change Detection Dataset}, 
  year={2021},
  volume={},
  number={},
  pages={1052-1061},
  doi={10.1109/CVPRW53098.2021.00116}}

@article{shu2025earthmind,
  title={Earthmind: Leveraging cross-sensor data for advanced earth observation interpretation with a unified multimodal llm},
  author={Shu, Yan and Ren, Bin and Xiong, Zhitong and Paudel, Danda Pani and Van Gool, Luc and Demir, Beg{\"u}m and Sebe, Nicu and Rota, Paolo},
  journal={arXiv preprint arXiv:2506.01667},
  year={2025}
}

@article{egybcd,
  title={AFDE-Net: Building change detection using attention-based feature differential enhancement for satellite imagery},
  author={Holail, Shimaa and Saleh, Tamer and Xiao, Xiongwu and Li, Deren},
  journal={IEEE Geoscience and Remote Sensing Letters},
  volume={20},
  pages={1--5},
  year={2023},
  publisher={IEEE}
}

@article{hrcuscd,
  title={AERNet: An attention-guided edge refinement network and a dataset for remote sensing building change detection},
  author={Zhang, Jindou and Shao, Zhenfeng and Ding, Qing and Huang, Xiao and Wang, Yu and Zhou, Xuechao and Li, Deren},
  journal={IEEE Transactions on Geoscience and Remote Sensing},
  volume={61},
  pages={1--16},
  year={2023},
  publisher={IEEE}
}

@article{levirmci,
  title={Change-agent: Toward interactive comprehensive remote sensing change interpretation and analysis},
  author={Liu, Chenyang and Chen, Keyan and Zhang, Haotian and Qi, Zipeng and Zou, Zhengxia and Shi, Zhenwei},
  journal={IEEE Transactions on Geoscience and Remote Sensing},
  volume={62},
  pages={1--16},
  year={2024},
  publisher={IEEE}
}

@article{tuecd,
  title={Building change detection in earthquake: A multiscale interaction network with offset calibration and a dataset},
  author={Liu, Yunlong and Zhang, Kai and Guan, Chunan and Zhang, Shanxin and Li, Hong and Wan, Wenbo and Sun, Jiande},
  journal={IEEE Transactions on Geoscience and Remote Sensing},
  volume={62},
  pages={1--17},
  year={2024},
  publisher={IEEE}
}

@article{optrsvg,
  title={Language-guided progressive attention for visual grounding in remote sensing images},
  author={Li, Ke and Wang, Di and Xu, Haojie and Zhong, Haodi and Wang, Cong},
  journal={IEEE Transactions on Geoscience and Remote Sensing},
  volume={62},
  pages={1--13},
  year={2024},
  publisher={IEEE}
}

@article{qwen3vl,
  title   = {Qwen3-VL Technical Report},
  author  = {Bai, Shuai and Cai, Yuxuan and Chen, Ruizhe and Chen, Keqin and Chen, Xionghui and Cheng, Zesen and Deng, Lianghao and Ding, Wei and Gao, Chang and Ge, Chunjiang and Ge, Wenbin and Guo, Zhifang and Huang, Qidong and Huang, Jie and Huang, Fei and Hui, Binyuan and Jiang, Shutong and Li, Zhaohai and Li, Mingsheng and Li, Mei and Li, Kaixin and Lin, Zicheng and Lin, Junyang and Liu, Xuejing and Liu, Jiawei and Liu, Chenglong and Liu, Yang and Liu, Dayiheng and Liu, Shixuan and Lu, Dunjie and Luo, Ruilin and Lv, Chenxu and Men, Rui and Meng, Lingchen and Ren, Xuancheng and Ren, Xingzhang and Song, Sibo and Sun, Yuchong and Tang, Jun and Tu, Jianhong and Wan, Jianqiang and Wang, Peng and Wang, Pengfei and Wang, Qiuyue and Wang, Yuxuan and Xie, Tianbao and Xu, Yiheng and Xu, Haiyang and Xu, Jin and Yang, Zhibo and Yang, Mingkun and Yang, Jianxin and Yang, An and Yu, Bowen and Zhang, Fei and Zhang, Hang and Zhang, Xi and Zheng, Bo and Zhong, Humen and Zhou, Jingren and Zhou, Fan and Zhou, Jing and Zhu, Yuanzhi and Zhu, Ke},
  journal = {arXiv preprint arXiv:2511.21631},
  year    = {2025}
}

@inproceedings{openrsd,
  title={OpenRSD: Towards open-prompts for object detection in remote sensing images},
  author={Huang, Ziyue and Feng, Yongchao and Liu, Ziqi and Yang, Shuai and Liu, Qingjie and Wang, Yunhong},
  booktitle={Proceedings of the IEEE/CVF International Conference on Computer Vision},
  pages={8384--8394},
  year={2025}
}

@article{layernorm,
  title={Layer normalization},
  author={Ba, Jimmy Lei and Kiros, Jamie Ryan and Hinton, Geoffrey E},
  journal={arXiv preprint arXiv:1607.06450},
  year={2016}
}

@article{park2025remote,
  title={Remote Sensing Large Vision-Language Model: Semantic-augmented Multi-level Alignment and Semantic-aware Expert Modeling},
  author={Park, Sungjune and Kim, Yeongyun and Kim, Se Yeon and Ro, Yong Man},
  journal={arXiv preprint arXiv:2506.21863},
  year={2025}
}

@inproceedings{objects365,
  title={Objects365: A large-scale, high-quality dataset for object detection},
  author={Shao, Shuai and Li, Zeming and Zhang, Tianyuan and Peng, Chao and Yu, Gang and Zhang, Xiangyu and Li, Jing and Sun, Jian},
  booktitle={Proceedings of the IEEE/CVF international conference on computer vision},
  pages={8430--8439},
  year={2019}
}

@inproceedings{gqa,
  title={Gqa: A new dataset for real-world visual reasoning and compositional question answering},
  author={Hudson, Drew A and Manning, Christopher D},
  booktitle={Proceedings of the IEEE/CVF conference on computer vision and pattern recognition},
  pages={6700--6709},
  year={2019}
}

@inproceedings{flickr30k,
  title={Flickr30k entities: Collecting region-to-phrase correspondences for richer image-to-sentence models},
  author={Plummer, Bryan A and Wang, Liwei and Cervantes, Chris M and Caicedo, Juan C and Hockenmaier, Julia and Lazebnik, Svetlana},
  booktitle={Proceedings of the IEEE international conference on computer vision},
  pages={2641--2649},
  year={2015}
}

@inproceedings{v3det,
  title={V3det: Vast vocabulary visual detection dataset},
  author={Wang, Jiaqi and Zhang, Pan and Chu, Tao and Cao, Yuhang and Zhou, Yujie and Wu, Tong and Wang, Bin and He, Conghui and Lin, Dahua},
  booktitle={Proceedings of the IEEE/CVF International Conference on Computer Vision},
  pages={19844--19854},
  year={2023}
}

@inproceedings{zero,
  title={Zero: Memory optimizations toward training trillion parameter models},
  author={Rajbhandari, Samyam and Rasley, Jeff and Ruwase, Olatunji and He, Yuxiong},
  booktitle={SC20: international conference for high performance computing, networking, storage and analysis},
  pages={1--16},
  year={2020},
  organization={IEEE}
}

@misc{qwen3.5,
    title  = {{Qwen3.5}: Towards Native Multimodal Agents},
    author = {{Qwen Team}},
    month  = {February},
    year   = {2026},
    url    = {https://qwen.ai/blog?id=qwen3.5}
}

@article{internvl3.5,
  title   = {InternVL3.5: Advancing Open-Source Multimodal Models in Versatility, Reasoning, and Efficiency},
  author  = {Wang, Weiyun and Gao, Zhangwei and Gu, Lixin and Pu, Hengjun and Cui, Long and Wei, Xingguang and Liu, Zhaoyang and Jing, Linglin and Ye, Shenglong and Shao, Jie and Wang, Zhaokai and Chen, Zhe and Zhang, Hongjie and Yang, Ganlin and Wang, Haomin and Wei, Qi and Yin, Jinhui and Li, Wenhao and Cui, Erfei and Chen, Guanzhou and Ding, Zichen and Tian, Changyao and Wu, Zhenyu and Xie, Jingjing and Li, Zehao and Yang, Bowen and Duan, Yuchen and Wang, Xuehui and Hou, Zhi and Hao, Haoran and Zhang, Tianyi and Li, Songze and Zhao, Xiangyu and Duan, Haodong and Deng, Nianchen and Fu, Bin and He, Yinan and Wang, Yi and He, Conghui and Shi, Botian and He, Junjun and Xiong, Yingtong and Lv, Han and Wu, Lijun and Shao, Wenqi and Zhang, Kaipeng and Deng, Huipeng and Qi, Biqing and Ge, Jiaye and Guo, Qipeng and Zhang, Wenwei and Zhang, Songyang and Cao, Maosong and Lin, Junyao and Tang, Kexian and Gao, Jianfei and Huang, Haian and Gu, Yuzhe and Lyu, Chengqi and Tang, Huanze and Wang, Rui and Lv, Haijun and Ouyang, Wanli and Wang, Limin and Dou, Min and Zhu, Xizhou and Lu, Tong and Lin, Dahua and Dai, Jifeng and Su, Weijie and Zhou, Bowen and Chen, Kai and Qiao, Yu and Wang, Wenhai and Luo, Gen},
  journal = {arXiv preprint arXiv:2508.18265},
  year    = {2025}
}

@inproceedings{kosmos2,
  title={Grounding multimodal large language models to the world},
  author={Peng, Zhiliang and Wang, Wenhui and Dong, Li and Hao, Yaru and Huang, Shaohan and Ma, Shuming and Ye, Qixiang and Wei, Furu},
  booktitle={International Conference on Learning Representations},
  year={2024}
}

\appendix

\section{Region Proposal Module Details}
\label{sec:app_rpm}

\paragraph{Detection Backbone.}
We use MM-Grounding-DINO~\cite{mmgroundingdino} with a Swin-Tiny backbone as the text-conditioned detection backbone, initialized from pretrained weights on Objects365~\cite{objects365}, GoldG~\cite{gqa,flickr30k}, Grit~\cite{kosmos2}, and V3Det~\cite{v3det}.
We fine-tune the detection head on five RS datasets (xView~\cite{xview}, DIOR~\cite{dior}, DOTA-v2.0~\cite{dota}, FAIR1M~\cite{fair1m}, SODA-A~\cite{soda}) for 20 epochs with a learning rate of $5{\times}10^{-5}$ and a step decay at epoch 15.
The backbone and language model are frozen during this fine-tuning; only the detection stack is updated.
Training uses 8 NVIDIA GeForce RTX 3090 GPUs with per-GPU batch size 1 and gradient accumulation of 4, yielding an effective batch size of 32.
We select the epoch-15 checkpoint based on zero-shot recall evaluation on held-out RS detection benchmarks, conducted over the unified class vocabulary described in Appendix~\ref{sec:app_classes}.

\paragraph{Prompt-Aware Class Filtering.}
At inference, we encode the user query and all 36 class names using all-MiniLM-L6-v2 \cite{minilm,sentencebert} and compute cosine similarity between the query embedding and each class name embedding.
The goal is to determine whether the query targets specific object categories (e.g., ``Identify all damaged buildings'' relates to \textit{building}) or is too general to narrow down (e.g., ``Describe the changes in this area'').
When the similarity distribution is sharply peaked toward a few classes, we use only those top-ranked classes as input to the detector; otherwise, all 36 classes are used.
Concretely, the filtering activates when three conditions are jointly met: (1) the highest cosine similarity score exceeds 0.35, (2) the z-score of that top score (relative to the mean and standard deviation of all 36 scores) exceeds 2.4, and (3) the standard deviation of the scores exceeds 0.06.
These thresholds were tuned on a small validation set to balance between reducing irrelevant proposals and avoiding missing relevant categories.
Across seven one-at-a-time threshold settings, AR@100 varied by at most 0.9 points on DIOR-RSVG and remained unchanged on S2Looking, and the filter retained the ground-truth category in 100\% of activated building-centric queries and 98.9\% of activated DIOR-RSVG queries.

\paragraph{Box Filtering.}
Raw proposals with detection confidence below 0.1 are discarded.
If the user query references a specific region (e.g., via a bounding box annotation), the corresponding reference box is added to the candidate set with an elevated confidence score to ensure it survives subsequent filtering.
Hard NMS with an IoU threshold of 0.5 is then applied to remove duplicate detections.
The final candidate set is capped at 100 regions per sample.
Sweeping this cap over 10, 25, 50, and 100 candidates showed that performance stabilizes beyond 50, with at most a 1.1-point variation.
Reference boxes are placed at the front of the candidate list (\roi{1}, \roi{2}, \ldots), and the remaining candidates are shuffled per training epoch to prevent the model from learning position-dependent biases.

\paragraph{ROI Exposure Policy.}
Not all tasks require the full set of candidate regions.
For localization and grounding tasks (e.g., change detection localization, visual grounding), the complete candidate set is provided to the LLM, allowing it to select among all proposals.
For region-level QA or captioning, where the user query refers to a specific region (e.g., ``Describe the building at \roi{3}''), only the referenced region is included, reducing input length and focusing the model's attention.
For scene-level tasks such as classification or general QA, no region tokens are provided at all, and the model operates purely on image patch tokens.
This flexible policy allows the same framework to handle diverse task types without architectural changes.

\section{Training Details}
\label{sec:app_training}

Our base model is Ovis2.5 \cite{ovis}, an open-source MLLM that integrates a native-resolution vision transformer (NaViT\cite{navit}) initialized from SigLIP2 \cite{siglip2} weights with a Qwen3 \cite{qwen3} language model backbone.
Ovis2.5 is pretrained through a multi-phase curriculum on large-scale multimodal data, providing strong general-purpose vision-language capabilities as our starting point.

All stages are trained on 8$\times$ NVIDIA A6000 GPUs using DeepSpeed ZeRO\cite{zero}, bf16 precision, and gradient checkpointing.
LoRA \cite{lora} is applied with rank 64, $\alpha{=}128$, and dropout 0.05 to all attention and feed-forward projections.
Region visual features are extracted from NaViT layers \{6, 13, 19, 26\} with $2{\times}2$ ROIAlign spatial bins.
The learnable scalars $\alpha_g$ and $\alpha_t$ are initialized near zero.
Maximum image resolution is $1024{\times}1024$ (fMoW: $512{\times}512$ due to memory), and maximum sequence length is 6000 tokens.

\textbf{Stage~1} trains the LLM via LoRA alongside the visual tokenizer head and the last vision encoder block (lr $2{\times}10^{-5}$; ViT last block lr $5{\times}10^{-6}$) for 2 epochs with per-GPU batch size 2 and gradient accumulation 4.

\textbf{Stage~2} is split into two phases with the LLM frozen throughout.
Phase~1 initializes and trains only the region projector (lr $1{\times}10^{-4}$) and ROI token embeddings (lr $2{\times}10^{-5}$).
Phase~2 activates the remaining region components including the geometry MLP, temporal embeddings, LayerNorm layers, and $\alpha_g$/$\alpha_t$ (lr $5{\times}10^{-5}$), expanding to multi-temporal and multi-target tasks with per-GPU batch size 2 and gradient accumulation 4.

\textbf{Stage~3} jointly trains the LLM via LoRA (lr $2{\times}10^{-5}$) and all region components (lr $5{\times}10^{-5}$) for 2 epochs on the full multi-task mixture, with per-GPU batch size 2 and gradient accumulation 4.
Gradient masking restricts embedding and output projection updates to ROI-specific token rows.

\section{Union Class Set for Region Proposer}
\label{sec:app_classes}

See Table~\ref{tab:union_classes} for the 36-category union class set.

\begin{table*}[t]
\centering
\small
\setlength{\tabcolsep}{4pt}
\begin{tabular}{@{}rlccccc@{}}
\toprule
\# & \textbf{Category} & \textbf{DIOR} & \textbf{DOTA-v2.0} & \textbf{FAIR1M} & \textbf{SODA-A} & \textbf{xView} \\
\midrule
1 & airplane & 1.9k & 19.8k & 32.1k & 40.6k & 1.9k \\
2 & helicopter & -- & 1.2k & -- & 1.7k & 0.1k \\
3 & airport & 0.7k & 0.4k & -- & -- & -- \\
4 & helipad & -- & 0.2k & -- & -- & 0.2k \\
5 & aircraft hangar & -- & -- & -- & -- & 0.3k \\
6 & ship & 27.3k & 106.7k & 41.9k & 85.6k & 8.1k \\
7 & harbor & 2.4k & 16.6k & -- & -- & -- \\
8 & container crane & -- & 0.5k & -- & -- & 0.2k \\
9 & container & -- & -- & -- & 175.2k & 2.5k \\
10 & container yard & -- & -- & -- & -- & 3.7k \\
11 & road vehicle & 13.7k & 429.0k & 290.4k & 668.5k & 441.8k \\
12 & rail vehicle & -- & -- & -- & -- & 7.0k \\
13 & train station & 0.5k & -- & -- & -- & -- \\
14 & constr. equip. & -- & -- & 26.9k & -- & 7.8k \\
15 & vehicle yard & -- & -- & -- & -- & 6.5k \\
16 & building & -- & -- & -- & -- & 546.4k \\
17 & constr. site & -- & -- & -- & -- & 1.7k \\
18 & storage tank & 3.1k & 17.6k & -- & 45.4k & 2.9k \\
19 & chimney & 0.6k & -- & -- & -- & -- \\
20 & wind turbine & 2.4k & -- & -- & 33.0k & -- \\
21 & utility tower & -- & -- & -- & -- & 0.9k \\
22 & dam & 0.5k & -- & -- & -- & -- \\
23 & bridge & 1.4k & 4.9k & 1.2k & -- & -- \\
24 & overpass & 1.3k & -- & -- & -- & -- \\
25 & roundabout & -- & 1.4k & 0.6k & -- & -- \\
26 & road intersect. & -- & -- & 7.0k & -- & -- \\
27 & toll station & 0.6k & -- & -- & -- & -- \\
28 & service area & 1.1k & -- & -- & -- & -- \\
29 & baseball field & 2.4k & 1.6k & 1.1k & -- & -- \\
30 & basketball court & 1.1k & 1.3k & 1.3k & -- & -- \\
31 & soccer field & -- & 0.9k & 0.9k & -- & -- \\
32 & tennis court & 4.9k & 6.6k & 2.9k & -- & -- \\
33 & running track & 1.2k & 1.0k & -- & -- & -- \\
34 & swimming pool & -- & 5.5k & -- & 37.8k & -- \\
35 & golf course & 0.5k & -- & -- & -- & -- \\
36 & stadium & 0.6k & -- & -- & -- & -- \\
\midrule
& \textbf{Total images} & 11.7k & 31.3k & 22.9k & 39.2k & 14.0k \\
& \textbf{Total instances} & 68.0k & 615.2k & 406.2k & 1087.8k & 1032.1k \\
\bottomrule
\end{tabular}
\caption{Union class set for the region proposer. Since each RS detection dataset defines its own class taxonomy, we construct a unified 36-category set by merging all source vocabularies and mapping each dataset's original classes to this shared space. Instance counts are shown per dataset; ``--'' indicates the category is absent.}
\label{tab:union_classes}
\end{table*}

\section{Dataset Summary}
\label{sec:app_data}

Table~\ref{tab:dataset_summary} summarizes training and evaluation data.

\begin{table*}[t]
\centering
\small
\setlength{\tabcolsep}{4pt}
\begin{tabular}{@{}llccccl@{}}
\toprule
\textbf{Source} & \textbf{Task} & \textbf{\# Train} & \textbf{\# Eval} & \textbf{\# Imgs} & \textbf{Metric} & \textbf{Stage} \\
\midrule
\multicolumn{7}{l}{\textit{TEOChatlas \cite{teochat}}} \\
\quad fMoW \cite{fmow} & Scene classification & 12,006 & 12,006 & 1--8 & Acc & 1, 3 \\
\quad xBD \cite{xbd} (LOC) & Building localization & 7,988 & 2,720 & 2 & F1 & 2, 3 \\
\quad xBD (CDC) & Damage classification & 10,560 & 3,600 & 2 & F1 & 2, 3 \\
\quad xBD (SRE) & Spatial referring expr. & 4,684 & 1,596 & 2 & F1 & 2, 3 \\
\quad xBD (QA/RQA) & Change QA / Region QA & 29,188 & 9,984 & 2 & Acc & 1--3 \\
\quad S2Looking \cite{s2looking} (CDL) & Change localization & 4,556 & 1,668 & 2 & F1 & 2, 3 \\
\quad S2Looking (SRE) & Spatial referring expr. & 2,788 & 1,020 & 2 & F1 & 2, 3 \\
\quad S2Looking (QA/RQA) & Change QA / Region QA & 13,676 & 5,012 & 2 & Acc & 1--3 \\
\quad QFabric \cite{qfabric} & Region/temporal QA & 36,552 & 7,636 & 2--5 & Acc/F1 & 2, 3 \\
\midrule
GeoChat Instruct \cite{geochat} & RS conversation & 296,097 & -- & 1 & -- & 1--3 \\
FIT-RS \cite{fitrs} & Fine-grained RS & 197,202 & -- & 1 & -- & 3 \\
DIOR-RSVG \cite{diorrsvg} & Visual grounding & 17,402 & 7,422 & 1 & Acc@0.5 & 2, 3 \\
OPT-RSVG \cite{optrsvg} & Visual grounding & 56,455 & -- & 1 & -- & 2, 3 \\
\midrule
LEVIR-MCI \cite{levirmci} & Change detection & 7,544 & 1,929 & 2 & F1 & 2, 3 \\
HRCUS-CD \cite{hrcuscd}$^\dag$ & Change detection & -- & 372 & 2 & F1 & -- \\
Additional CD & Change detection & $\sim$20k & -- & 2 & -- & 2, 3 \\
\bottomrule
\end{tabular}
\caption{Dataset summary. $^\dag$Zero-shot evaluation only.}
\label{tab:dataset_summary}
\end{table*}

\begin{table*}[t]
\centering
\small
\setlength{\tabcolsep}{3pt}
\begin{tabular}{@{}p{2.4cm}p{1.0cm}p{5.5cm}p{5.5cm}@{}}
\toprule
\textbf{Task} & \textbf{\# Imgs} & \textbf{Example Prompt} & \textbf{Example Response} \\
\midrule
\multicolumn{4}{l}{\textit{Evaluation tasks}} \\
\addlinespace
xBD LOC & 2 & Identify all the buildings in the first image. Include region token(s) for each identified region. & \roi{1}, \roi{5}, \roi{8}, \roi{12} \\
\addlinespace
S2Looking CDL & 2 & Identify all changed buildings. Include region token(s) for each identified region. If there are no such regions, do not include any region tokens. & \roi{2}, \roi{7}, \roi{15} \\
\addlinespace
xBD SRE & 2 & Identify the destroyed buildings in this area: \roi{3}. Include region token(s) for each identified region. & \roi{5}, \roi{9} \\
\addlinespace
xBD CDC & 2 & How severe is the damage to this building? \roi{5} & Major damage. \\
\addlinespace
xBD QA & 2 & Are there any destroyed buildings in the area? Answer with one word. & Yes. \\
\addlinespace
xBD RQA & 2 & How has this building \roi{3} changed between the first and second image? & The building has sustained major structural damage. \\
\addlinespace
QFabric RTQA & 2--5 & What change has occurred in \roi{2} between image 1 and image 3? & New construction has appeared. \\
\addlinespace
QFabric TRE & 2--5 & In which image was the area \roi{5} first developed? & Image 3. \\
\addlinespace
fMoW & 1--8 & What type of functional area is shown in this satellite image? & Airport. \\
\addlinespace
DIOR-RSVG & 1 & A small ship docked on the right side of the harbor. & \roi{15} \\
\addlinespace
LEVIR-MCI & 2 & Identify the changed building areas in these two satellite domain images. & \roi{30}, \roi{12}, \roi{31} \\
\midrule
\multicolumn{4}{l}{\textit{Training-only tasks}} \\
\addlinespace
GeoChat Instruct & 1 & Can you describe what you see in this satellite image? & The image shows a residential area with several low-rise buildings and tree cover along the streets. \\
\addlinespace
FIT-RS & 1 & Describe the objects and layout visible in this remote sensing image in detail. & A large industrial complex is visible in the center, surrounded by storage tanks and connected by roads to a nearby port facility. \\
\addlinespace
OPT-RSVG & 1 & The red-roofed building in the northeast corner of the residential block. & \roi{8} \\
\addlinespace
Region Captioning & 2 & Describe the building at \roi{3}. & The building at \roi{3} appears to be a two-story residential structure with a light-colored roof. \\
\addlinespace
Grounded Desc. & 2 & Describe how the buildings have changed. Include region token(s) for each identified region. & Buildings at \roi{2} and \roi{7} show significant structural damage, while the area at \roi{11} remains intact. \\
\bottomrule
\end{tabular}
\caption{Example prompts and responses for evaluation and training-only tasks. Region tokens (\roi{k}) refer to candidate regions from the region proposal module. In practice, prompts are preceded by image tokens and a system prefix describing available region tokens.}
\label{tab:task_examples}
\end{table*}

\section{Comparison with Recent Generalist MLLMs}
\label{app:generalist}

We additionally evaluate Qwen3.5-9B~\cite{qwen3.5} and InternVL3.5-8B~\cite{internvl3.5}, two recent open-weight generalist MLLMs of comparable parameter scale, under the same evaluation protocol and output adapters as in Section~4.1. Both models are evaluated in a zero-shot manner. Table~\ref{tab:generalist} reports results on the five localization benchmarks used for this comparison, together with the models from Table~\ref{tab:loc_results}. The stronger generalist models improve over Qwen3-VL, particularly on single-image grounding, but the gap to our model remains large on every benchmark, and change detection localization stays below 21 F1 for Qwen3.5-9B and 34 F1 for InternVL3.5-8B.

\begin{table}[t]
\centering\small
\setlength{\tabcolsep}{3.5pt}
\resizebox{\columnwidth}{!}{
\begin{tabular}{@{}lccccc@{}}
\toprule
 & \textbf{xBD} & \textbf{S2Looking} & \textbf{DIOR-RSVG} & \textbf{HRCUS} & \textbf{LEVIR} \\
\textbf{Model} & LOC & CDL & Acc@0.5 & F1 & F1 \\
\midrule
Qwen3-VL & 17.3$^*$ & 11.8$^*$ & 53.8$^*$ & 21.7$^*$ & 13.3$^*$ \\
Qwen3.5-9B & 22.8$^*$ & 12.5$^*$ & 58.3$^*$ & 20.8$^*$ & 21.2$^*$ \\
InternVL3.5-8B & 27.2$^*$ & 20.3$^*$ & 69.0$^*$ & 33.2$^*$ & 26.4$^*$ \\
TEOChat & 38.9 & 34.5 & 27.6 & 33.3$^*$ & 26.4$^*$ \\
EarthDial & 24.1 & 2.6$^*$ & 39.6 & 4.7$^*$ & 12.6$^*$ \\
Ovis2.5-FT & 32.3 & 36.4 & 69.5 & 54.4$^\dag$ & 56.4 \\
\textbf{Ours} & \textbf{69.4} & \textbf{50.5} & \textbf{77.3} & \textbf{55.0}$^\dag$ & \textbf{58.8} \\
\bottomrule
\end{tabular}%
}
\caption{Localization results (\%) on the five benchmarks evaluated for the generalist comparison. $^*$Zero-shot (not trained on this dataset). $^\dag$HRCUS-CD not in training data.}
\label{tab:generalist}
\end{table}

\section{Full Component Ablation}
\label{app:full_ablation}

Table~\ref{tab:full_ablation} extends the ablation in Table~\ref{tab:ablation} by separately removing the spatial and temporal cues, evaluated on the three benchmarks used in this analysis. Removing the spatial cue causes most of the degradation, with DIOR-RSVG falling from 77.3 to 33.5, while removing only the temporal cue leads to small but consistent drops on all three benchmarks.

\begin{table}[t]
\centering\small
\begin{tabular}{@{}lccc@{}}
\toprule
\textbf{Configuration} & \textbf{S2L CDL} & \textbf{DIOR-RSVG} & \textbf{LEVIR} \\
\midrule
Full model & \textbf{50.5} & \textbf{77.3} & \textbf{58.8} \\
w/o temp.\ cue & 50.1 & 76.1 & 58.0 \\
w/o spat.\ cue & 39.2 & 33.5 & 56.5 \\
w/o spat.+temp.\ cues & 39.0 & 32.3 & 56.6 \\
Text-coordinate selection & 47.5 & 74.7 & 55.7 \\
Coordinate generation & 36.4 & 69.5 & 56.4 \\
\bottomrule
\end{tabular}
\caption{Component ablation over all six configurations. Text-coordinate selection replaces region feature tokens with textual coordinates of the candidates, and coordinate generation is the No ROI baseline from Table~\ref{tab:ablation}.}
\label{tab:full_ablation}
\end{table}

\section{Inference Efficiency}
\label{app:efficiency}

Table~\ref{tab:efficiency} compares the end-to-end inference cost of region selection and coordinate generation under identical hardware and batch settings. Region selection shortens the generated output from 39.1 to 6.4 tokens per sample on S2Looking and from 26.0 to 3.4 on DIOR-RSVG. The region proposal module adds 0.11 to 0.22 seconds per sample, and the injected region tokens add prefill rather than decoding cost. As a result, total wall-clock time drops by a factor of 4.7 to 5.3.

\begin{table}[t]
\centering\small
\setlength{\tabcolsep}{4pt}
\begin{tabular}{@{}lcccc@{}}
\toprule
 & \multicolumn{2}{c}{\textbf{S2Looking}} & \multicolumn{2}{c}{\textbf{DIOR-RSVG}} \\
\cmidrule(lr){2-3}\cmidrule(lr){4-5}
 & Coord. & Region & Coord. & Region \\
\midrule
Output tokens / sample & 39.1 & 6.4 & 26.0 & 3.4 \\
Total time (s / sample) & 6.75 & \textbf{1.27} & 4.18 & \textbf{0.89} \\
\bottomrule
\end{tabular}
\caption{Inference efficiency of coordinate generation (Coord.) and region selection (Region). Total time includes the region proposal module, which adds 0.11 to 0.22 seconds per sample.}
\label{tab:efficiency}
\end{table}

\section{Qualitative Examples}
\label{sec:app_qual}

\begin{figure*}[t]
\centering
\includegraphics[width=\textwidth]{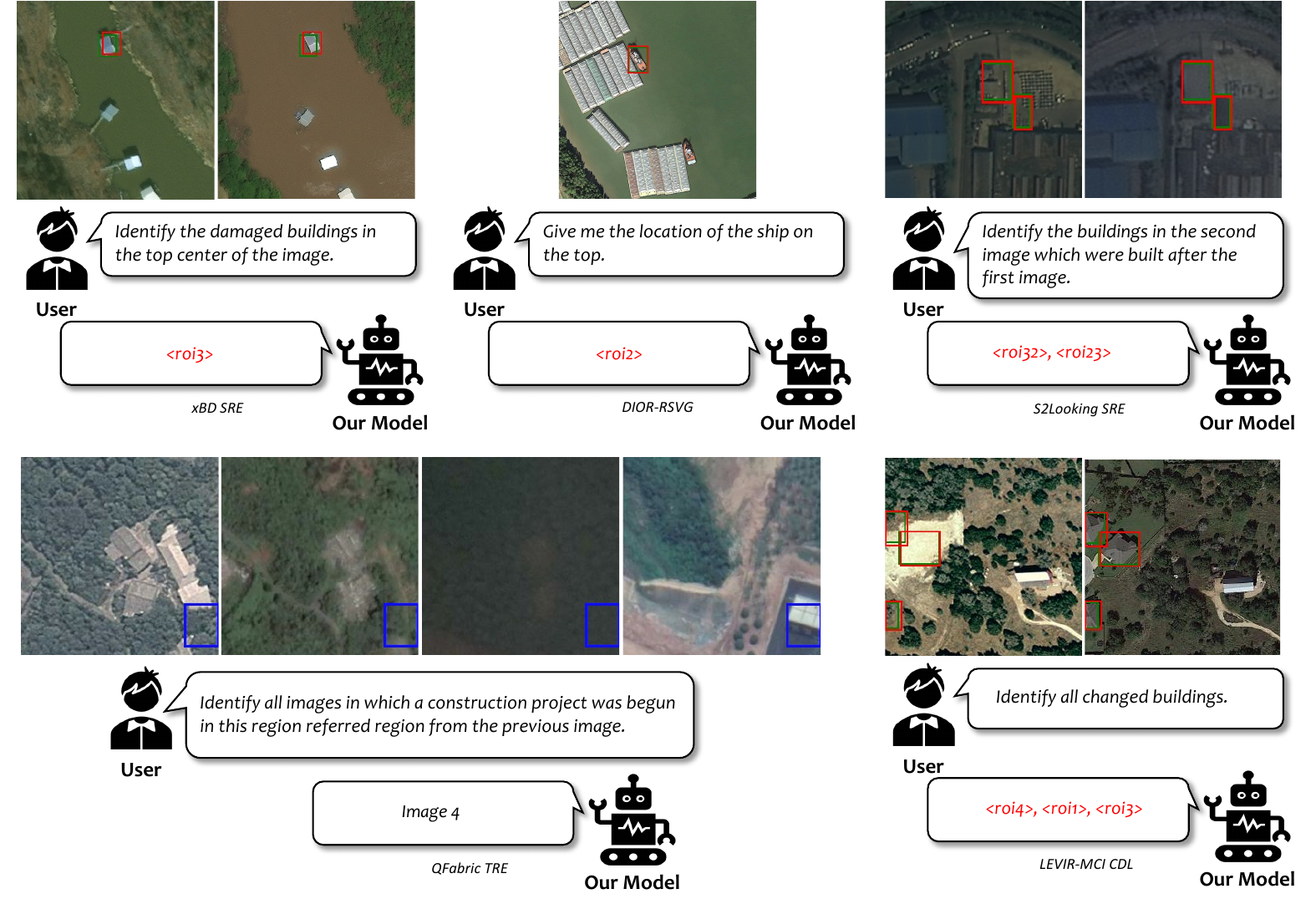}
\caption{Qualitative examples across five tasks. \textbf{Top row:} xBD SRE (bi-temporal spatial referring expression), DIOR-RSVG (single-image visual grounding), and S2Looking SRE (bi-temporal spatial referring expression). \textbf{Bottom row:} QFabric TRE (temporal referring expression over four images) and LEVIR-MCI CDL (bi-temporal change detection localization). In all cases, the model produces region token selections that correctly correspond to the queried targets, and identifies the correct temporal frame in which a change occurred across a multi-image sequence.}
\label{fig:app_qual}
\end{figure*}

Figure~\ref{fig:app_qual} presents qualitative examples of our model across five tasks.
For spatial referring expression (xBD SRE, S2Looking SRE), the model correctly selects the regions matching spatial and semantic constraints from the query.
For single-image visual grounding (DIOR-RSVG), it identifies the described object through region token selection.
For temporal referring expression (QFabric TRE), the model identifies the correct temporal frame in which a change occurred across a multi-image sequence.
For change detection localization (LEVIR-MCI CDL), it selects multiple changed building regions from a bi-temporal image pair.

\end{document}